\documentclass[runningheads]{llncs}

\usepackage{multirow}
\usepackage{times}
\usepackage{paralist}
\usepackage{latexsym}
\usepackage{todonotes}
\usepackage{adjustbox}
\usepackage{subcaption}
\usepackage{tcolorbox}
\usepackage{pgfplots}
\pgfplotsset{compat=1.9}

\usepackage[T1]{fontenc}

\usepackage{microtype}
\usepackage[ruled,vlined]{algorithm2e}

\usepackage{graphicx}
\usepackage{microtype}

\usepackage{booktabs} 
\usepackage{float}
\usepackage{listings}
\usepackage{tikz}
\usetikzlibrary{patterns}
\usetikzlibrary {shapes.misc}
\usetikzlibrary{shapes.geometric,arrows,positioning,fit}
\usepackage[font=small,skip=6pt]{caption}
\usepackage{hyperref}
\usepackage{xspace}

\usepackage{pifont}
\usepackage{amsmath}
\usepackage{mathtools}
\usepackage{amsfonts}
\usepackage{verbatimbox}

\usepackage{comment}

\usepackage[nolist]{acronym}

\begin{document}

\title{Learning to Retrieve with Weakened Labels: Robust Training under Label Noise}

\titlerunning{Robust Retrieval under Noise}

\author{Arnab Sharma}
\institute{
Heinz Nixdorf Institute, Paderborn University, Germany\\
\email{\{arnab.sharma\}@uni-paderborn.de}}

\maketitle

\begin{abstract}
Neural Encoders are frequently used in the NLP domain to perform dense retrieval tasks, for instance, to generate the candidate documents for a given query in question-answering tasks. However, sparse annotation and label noise in the training data make it challenging for training or fine-tuning such retrieval models. 
Although existing works have attempted to mitigate these problems by incorporating modified loss functions or data-cleaning, these approaches either require some hyperparameters to tune during training or add substantial complexity to the training setup. In this work, we consider a {\em label weakening} approach to generate robust retrieval models in the presence of label noise. Instead of enforcing a single, potentially erroneous label for each query–document pair, we allow for a set of plausible labels derived from both the observed supervision and the model’s confidence scores.  We perform an extensive evaluation considering two retrieval models, one re-ranking model, considering four diverse ranking datasets. To this end, we also consider a realistic noisy setting by using a {\em semantic-aware} noise generation technique to generate different ratios of noise. 
Our initial results show that label weakening can improve the performance of the retrieval tasks in comparison to 10 different state-of-the-art loss functions.


\end{abstract}

\section{Introduction}

Neural Encoders play an important role in information retrieval tasks, such as in question-answering and entity ranking, amongst others. In recent years, with the advent of transformer-based architecture, language models such as BERT~\cite{DevlinCLT19,gupta2024} are frequently being used as encoders, achieving state-of-the-art performances. These models typically work by generating context-aware vector representations of queries and candidates, and then computing similarity scores through inner products or other distance measures~\cite{ReimersG19,abs-1901-04085,abs-2403-10407,ZhangHS22,WuPJRZ20}. The resulting scores are subsequently used either to retrieve the most relevant candidates from a large collection (dense retrieval) or to re-rank a smaller set of pre-retrieved candidates (re-ranking)~\cite{ReimersG19,abs-1901-04085,abs-2403-10407,ZhangHS22,WuPJRZ20}. 
Despite their effectiveness, such models often face challenges arising from label noise, sparse supervision, and distribution shifts. This necessitates the training of robust neural encoder models.

In the context of retrieval and ranking, label noise is prevalent in real-world datasets, since those are often generated through human annotation~\cite{ZerveasRCE22,ZerveasRE23}. This involves subjective judgments of relevance, inconsistencies across annotators, and mislabeling of relevant or non-relevant documents, all of which can adversely affect the training instances and lead to degraded model performance. Note that although in the general learning setting and in NLP domain, this has been studied extensively~\cite{LiuT16,NatarajanDRT13,ChengV22,GhoshSG22,LiuAGD22}, training retrieval models under label noise remains relatively underexplored~\cite{ZerveasRE23,PenhaH21}. 

To mitigate this issue, in this work, we propose to incorporate a label weakening  strategy considering the idea of {\em superset learning}~\cite{HullermeierC15} and {\em data ambiguation}~\cite{LienenH24}. The main idea herein is that instead of committing to potentially corrupted binary relevance labels, we construct {\em ambiguated target sets}. These retain the original annotation while augmenting it with additional plausible labels inferred from model predictions. Most importantly, unlike existing approaches for handling noisy labels in NLP models, the label weakening approach does not attempt to directly correct or denoise individual annotations. Instead, it weakens the supervision signal by broadening the set of acceptable targets.
Therefore, representing supervision through the sets of candidate labels, this approach reduces the risk of memorizing noisy annotations, maintains flexibility in early training, and allows the model to disambiguate relevance information in a robust, data-driven manner.  
We perform an extensive evaluation comparing the label weakening approach to several robust loss functions, considering two retrieval models trained on 4 different datasets. 
Experimental results suggest the effectiveness of the label weakening approach.
In summary, in this paper, we make the following contributions.

\begin{itemize}
    \item We introduce a label weakening strategy for retrieval models under label noise.
    \item We formalize the use of ambiguated target sets in retrieval. 
    \item We provide a comprehensive experimental evaluation of the proposed method~\footnote{The code and accompanying documentation required to reproduce the results presented in this paper are available at:~\url{https://doi.org/10.5281/zenodo.17258776}}.
\end{itemize}

\section{Related Work}

In recent years, transformer-based bi-encoders, such as BERT-based dense retrievers, have emerged as one of the most prominent choices for retrieval and ranking tasks~\cite{ReimersG19,abs-1901-04085,abs-2403-10407}. To this end, Reimers et al.~\cite{ReimersG19} proposed Sentence-BERT, which works as a neural encoder to generate the embeddings for the queries and the candidate documents. There are some extensions of this work in passage re-ranking tasks~\cite{abs-1901-04085} and by Wang et al.~\cite{abs-2212-03533} in developing E5 models generating semantically rich sentence embeddings. Furthermore, Dêjan et al.~\cite{abs-2403-10407} proposed a cross-encoder that achieves state-of-the-art performance in ranking, however, at a high cost. 
Note that although significant progress has been made to develop effective retrieval and ranking approaches, their performance critically depends on the quality of supervision. 
For instance,~\cite{ZerveasRE23} argued that datasets like MS MARCO, while large-scale, contain sparse judgments and are prone to false negatives, i.e., passages that are marked non-relevant despite being semantically useful. To alleviate this,~\cite{PenhaH21} investigated the use of label smoothing~\cite{SzegedyVISW16} in this context. More specifically, they proposed an adaptive version of label smoothing called weakly supervised label smoothing, which uses retrieval scores of negatives as a weak supervision signal. This means that negatives more similar to the query are assigned a higher probability mass (instead of being treated as hard 0 labels).
Zerveas et al.~\cite{ZerveasRCE22} have further explored list-wise training objectives that consider entire candidate sets instead of isolated pairs, enabling models to capture inter-candidate relations more effectively. In a later work, they~\cite{ZerveasRE23} proposed {\em reciprocal nearest neighbor} approach to mitigate the issue of sparse annotation in the question-answering datasets. To this end, they proposed {\em evidence-based} smoothing which leverages structural similarity between candidates and the ground-truth passages using both geometric similarity and reciprocal nearest neighbor-based Jaccard measures.
These approaches primarily employ smoothing-based loss functions to enhance the performance of retrieval models in general. However, whether these approaches can be used for dealing with noisy labels is not explored. Additionally, these approaches require tuning the hyperparameters to get the best results and or strong assumptions on noise distributions. 

Our label-weakening approach is based on the idea of superset learning from~\cite{HullermeierC15,LienenH24} wherein a convex superset loss is optimized via a possibility distribution. 
Additionally, it reduces memorization of noisy annotations while preserving flexibility for the model to disambiguate relevance during training. Note that recent works in NLP have increasingly focused on developing robust models to label noise. For example, instance-adaptive training by Jin et al.~\cite{JinSXY21} dynamically predicts per-example robustness parameters for noise-robust losses. Other approaches include noise modeling~\cite{GargRT21}, meta-rectification networks~\cite{SunGWHY22}, instance–label pair correction in multi-label settings~\cite{xu-etal-2024-noisy}. 

\section{Preliminaries}

We start by giving a formal definition of label weakening approach in the retrieval and ranking setup.
Consider $\mathcal{C}$ denotes the  entire set of documents. 
Given a query $q \in \mathcal{Q}$, the candidate retrieval stage 
defines a mapping $\mathcal{R} : \mathcal{Q} \times \mathcal{C} \;\rightarrow\; 2^{\mathcal{C}}$, such that $\mathcal{D}_q = \mathcal{R}(q, \mathcal{C})$, and  $\mathcal{D}_q \subseteq \mathcal{C}$, where 
$|\mathcal{D}_q| = K \ll |\mathcal{C}|.$
Here, $\mathcal{R}$ returns a reduced set of $K$ candidates $\mathcal{D}_q$ for each query $q$.  
The central requirement of candidate retrieval is that the set of relevant documents $\mathcal{R}_q \subseteq \mathcal{C}$ is, with high probability, preserved within the retrieved candidate set as $\mathcal{R}_q \cap \mathcal{D}_q \neq \emptyset, 
 \text{ and ideally } \mathcal{R}_q \subseteq \mathcal{D}_q.$
In practice, retrieval is implemented via a scoring function $s(q,d)$, which assigns a real-valued relevance score to each document $d \in \mathcal{C}$ with respect to $q$. The candidate set is then obtained by selecting the top-$K$ highest-scoring documents, $
\mathcal{D}_q = \operatorname{TopK}_{d \in \mathcal{C}} \, s(q,d).$

In the retrieval setting, label noise arises when the supervision for a query $q$ does not reliably indicate which candidates in $\mathcal{D}_q$ are truly relevant. 
To address this, we adopt the label weakening approach, inspired by data ambiguation~\cite{LienenH24}, wherein single-valued supervision is replaced with set-valued targets. More specifically, instead of committing to a single relevant document $d^+$, the learner is provided with an ambiguated relevant set $\mathcal{R}^\ast_q \supseteq \mathcal{R}_q$.  $\mathcal{R}^\ast_q$ contains both the annotated relevant candidates and additional documents deemed {\em plausibly relevant}. 
The set $\mathcal{R}^\ast_q$ is obtained by a confidence-thresholding strategy, given model predictions $p(\cdot \mid q)$ over $\mathcal{D}_q$, all candidates whose predicted relevance exceeds a threshold $\beta$ are included in $\mathcal{R}^\ast_q$.
Formally, we define for each query--candidate pair $(q,d)$ a possibility distribution $\pi_{q}(d)$ as follows.
\[
\pi_q(d) =
\begin{cases}
1 & \text{if } d \in \mathcal{R}_q \ \text{or}\ p(d \mid q) \geq \beta, \\
\alpha & \text{otherwise},
\end{cases}
\]
where $\beta \in [0,1]$ is a confidence threshold and $\alpha \in [0,1)$ is a relaxation parameter.  
The induced credal set $\mathcal{Q}_{q,d}$ then represents all probabilistic relevance distributions consistent with $\pi_q$, thereby weakening supervision and reducing the risk of memorizing noisy labels. Essentially, training with label weakening corresponds to minimizing the optimistic superset loss as proposed by~\cite{HullermeierC15}, 
This formulation allows the model to select the most favorable interpretation among the plausibly relevant candidates, while still being guided by the original supervision. 
Given the ambiguated relevant set $\mathcal{R}^\ast_q$ and its complement $\mathcal{N}^\ast_q = \mathcal{D}_q \setminus \mathcal{R}^\ast_q$, we define
\[
\mathcal{L}_{\text{amb}}(q) 
= \min_{d^+ \in \mathcal{R}^\ast_q} 
\; \sum_{d^- \in \mathcal{N}^\ast_q} 
\ell\!\left(f(q,d^+), f(q,d^-)\right),
\]
where $\ell(\cdot,\cdot)$ is a pairwise ranking loss (e.g., logistic or margin-based).  
This formulation allows the ranker to select the most plausible positive instance from $\mathcal{R}^\ast_q$, while still contrasting it against non-relevant candidates. 
Thus, instead of memorizing potentially corrupted labels, the ranking model optimistically exploits ambiguity to retain discriminative supervision, improving robustness against label noise in ranking tasks.
After generating the list of candidate documents $\mathcal{D}_q$, the ranking stage assigns 
a relevance score to each candidate. For this a scoring function is defined as, $f : \mathcal{Q} \times \mathcal{C} \;\rightarrow\; \mathbb{R}$. Herein, each candidate $d \in \mathcal{D}_q$ obtains a score $f(q,d)$. 
The final ranking is obtained by sorting the candidates as $\pi_q = \operatorname{argsort}_{d \in \mathcal{D}_q} \; f(q,d)$,
where $\pi_q$ denotes the permutation of candidates in descending order of predicted relevance.

Note that, although this idea might read similar to the approach of {\em self-labeling} due to the generation of ambiguity set $R_q^*$ using $p(d|q)$, label weakening is conceptually different. In the former, new pseudo-labels replace the supervision signal and are treated as additional positives. However, in contrast, label weakening preserves all original labels and only bounds the supervision by a possibility distribution $\pi_q(d)$. 
Moreover, to avoid a potential feedback loop where the model could reinforce its own confident mistakes in early training stages, we perform the following two operations. 

\noindent
\textit{(i) Delayed ambiguity update:} The confidence scores used to form ambiguated targets are derived from the current epoch’s predictions, which are recomputed after each full pass through the data. This avoids runaway reinforcement from single-batch noise.

\noindent
\textit{(ii) Fixed $\alpha < 1$ and high $\beta$:} Early predictions below $\beta$ are not incorporated, meaning low-confidence or incorrect candidates are excluded. Thus, the method functions more like a soft consistency regularizer than a self-labeling mechanism.

\setlength{\tabcolsep}{4pt} 
\begin{table*}[!t]
    \centering
    \caption{Recall scores across varying noise ratios, datasets, and loss functions for the \textbf{dual bi-encoder} model. Here, 
\textbf{CE}: Cross Entropy; 
\textbf{NCE}: Normalized Cross Entropy; 
\textbf{GCE}: Generalized Cross Entropy; 
\textbf{AGCE}: Active Generalized Cross Entropy; 
\textbf{AUL}: Adaptive Unhinged Estimator; 
\textbf{LS}: Label Smoothing; 
\textbf{ELS}: Evidence-based Label Smoothing; 
\textbf{NAGCE}: Combined NCE and AGCE; 
\textbf{NAUL}: Combined NCE and AUL; 
\textbf{LR}: Label Relaxation; 
\textbf{LW} (ours): Label Weakening.}
    \label{tab:combined-recall}
    \resizebox{\textwidth}{!}{%
    \begin{tabular}{ll|ccccccccccc}
        \toprule
        \textbf{Dataset} & \textbf{Noise} 
        & \textbf{CE} & \textbf{NCE} & \textbf{GCE} & \textbf{AGCE} & \textbf{AUL}
        & \textbf{LS} & \textbf{ELS} & \textbf{NAGCE} & \textbf{NAUL} & \textbf{LR} & \textbf{LW (ours)} \\
        \midrule
        \multicolumn{13}{c}{\textbf{Dual Bi-Encoder Model}} \\
        \midrule
        MS MARCO & 0.0 & 0.888 & 0.848 & 0.906 & 0.899 & 0.903 & 0.900 & 0.907 & 0.884 & \textbf{0.909} & \textbf{0.938} & 0.896 \\
                 & 0.1 & 0.577 & 0.839 & 0.892 & 0.898 & \textbf{0.906} & 0.899 & 0.901 & 0.615 & 0.902 & \textbf{0.934} & 0.895 \\
                 & 0.2 & 0.333 & 0.779 & 0.882 & 0.892 & \textbf{0.903} & 0.893 & 0.897 & 0.454 & 0.805 & \textbf{0.930} & 0.887 \\
                 & 0.3 & 0.347 & 0.726 & 0.760 & 0.881 & \textbf{0.890} & 0.882 & 0.883 & 0.269 & 0.869 & \textbf{0.888} & 0.888 \\
                 & 0.4 & 0.153 & 0.536 & 0.546 & 0.833 & 0.682 & 0.811 & 0.882 & 0.255 & 0.835 & 0.830 & \textbf{0.863} \\
                 & 0.5 & 0.151 & 0.408 & 0.343 & 0.629 & 0.463 & 0.632 & 0.582 & 0.189 & 0.466 & 0.811 & \textbf{0.812} \\
        \midrule
        LCQuAD  & 0.0 & \textbf{0.918} & 0.911 & 0.914 & 0.917 & 0.909 & 0.912 & 0.914 & 0.915 & 0.916 & 0.911 & 0.915 \\
                 & 0.1 & \textbf{0.913} & 0.904 & 0.912 & 0.910 & 0.907 & 0.909 & 0.905 & 0.914 & 0.905 & 0.905 & 0.910 \\
                 & 0.2 & 0.785 & 0.887 & 0.791 & 0.812 & 0.900 & 0.901 & 0.901 & 0.823 & 0.900 & 0.884 & \textbf{0.905} \\
                 & 0.3 & 0.754 & 0.856 & 0.764 & 0.760 & 0.896 & 0.796 & 0.799 & 0.761 & 0.882 & 0.850 & \textbf{0.906} \\
                 & 0.4 & 0.782 & 0.738 & 0.800 & 0.775 & 0.895 & 0.782 & 0.777 & 0.781 & 0.899 & 0.818 & \textbf{0.901} \\
                 & 0.5 & 0.790 & 0.700 & 0.772 & 0.833 & 0.890 & 0.771 & 0.782 & 0.830 & 0.891 & 0.790 & \textbf{0.900} \\
        \midrule
        Mintaka & 0.0 & 0.516 & 0.512 & 0.512 & 0.513 & 0.515 & 0.511 & 0.521 & 0.520 & 0.522 & 0.546 & \textbf{0.526} \\
                 & 0.1 & 0.494 & 0.504 & 0.518 & 0.512 & 0.500 & 0.501 & \textbf{0.523} & 0.517 & 0.509 & 0.543 & 0.521 \\
                 & 0.2 & 0.488 & 0.470 & 0.500 & 0.492 & 0.515 & 0.500 & \textbf{0.526} & 0.488 & 0.516 & 0.508 & 0.520 \\
                 & 0.3 & 0.487 & 0.445 & 0.460 & 0.484 & 0.501 & 0.491 & 0.506 & 0.481 & 0.503 & 0.470 & \textbf{0.507} \\
                 & 0.4 & 0.490 & 0.401 & 0.481 & 0.477 & 0.480 & 0.488 & 0.481 & 0.465 & 0.497 & 0.437 & \textbf{0.507} \\
                 & 0.5 & 0.466 & 0.399 & 0.448 & 0.482 & 0.498 & 0.482 & 0.472 & 0.468 & 0.481 & 0.401 & \textbf{0.503} \\
        \midrule
        AIDA    & 0.0 & \textbf{0.226} & 0.207 & 0.219 & 0.209 & 0.207 & 0.211 & 0.221 & 0.220 & 0.208 & 0.221 & 0.210 \\
                 & 0.1 & 0.210 & 0.198 & 0.217 & 0.204 & 0.216 & 0.202 & 0.212 & \textbf{0.219} & 0.199 & 0.200 & 0.203 \\
                 & 0.2 & 0.208 & 0.194 & 0.214 & 0.207 & 0.214 & 0.207 & 0.212 & 0.189 & \textbf{0.219} & 0.185 & 0.215 \\
                 & 0.3 & 0.194 & 0.179 & 0.214 & 0.194 & 0.210 & 0.214 & 0.205 & 0.195 & 0.202 & 0.178 & \textbf{0.219} \\
                 & 0.4 & 0.189 & 0.179 & \textbf{0.216} & 0.202 & 0.200 & 0.200 & 0.204 & 0.199 & 0.203 & 0.172 & 0.212 \\
                 & 0.5 & 0.184 & 0.165 & 0.196 & 0.186 & 0.199 & 0.190 & 0.202 & 0.195 & 0.193 & 0.178 & \textbf{0.212} \\
        \midrule
       
        \multicolumn{13}{c}{\textbf{Recall scores for E5 model}} \\
        \midrule
        MS MARCO & 0.0 & 0.842 & 0.733 & 0.836 & 0.861 & 0.867 & 0.967 & 0.968 & 0.805 & 0.854 & 0.964 & \textbf{0.969} \\
                 & 0.1 & 0.822 & 0.779 & 0.834 & 0.832 & 0.842 & 0.962 & 0.962 & 0.353 & 0.833 & \textbf{0.971} & 0.968 \\
                 & 0.2 & 0.788 & 0.744 & 0.815 & 0.784 & 0.796 & 0.952 & 0.950 & 0.147 & 0.798 & 0.961 & \textbf{0.964} \\
                 & 0.3 & 0.792 & 0.678 & 0.805 & 0.890 & 0.780 & 0.945 & 0.955 & 0.073 & 0.783 & \textbf{0.967} & 0.959 \\
                 & 0.4 & 0.800 & 0.557 & 0.791 & 0.834 & 0.789 & 0.944 & 0.945 & 0.053 & 0.806 & 0.941 & \textbf{0.943} \\
                 & 0.5 & 0.733 & 0.609 & 0.766 & 0.781 & 0.779 & 0.909 & 0.911 & 0.042 & 0.783 & 0.911 & \textbf{0.913} \\
        \midrule
        LCQuAD & 0.0 & 0.827 & 0.827 & 0.827 & 0.827 & 0.836 & 0.880 & 0.880 & 0.827 & 0.836 & 0.882 & \textbf{0.890} \\
                & 0.1 & 0.719 & 0.719 & 0.719 & 0.719 & 0.833 & 0.870 & 0.870 & 0.719 & 0.833 & \textbf{0.874} & 0.872 \\
                & 0.2 & 0.612 & 0.612 & 0.612 & 0.612 & 0.824 & 0.855 & 0.855 & 0.612 & 0.824 & 0.859 & \textbf{0.861} \\
                & 0.3 & 0.527 & 0.527 & 0.527 & 0.527 & 0.809 & 0.830 & 0.830 & 0.527 & 0.809 & 0.835 & \textbf{0.840} \\
                & 0.4 & 0.433 & 0.433 & 0.433 & 0.433 & 0.792 & 0.800 & 0.800 & 0.433 & 0.792 & 0.806 & \textbf{0.812} \\
                & 0.5 & 0.357 & 0.357 & 0.357 & 0.357 & 0.783 & 0.772 & 0.772 & 0.357 & 0.783 & 0.786 & \textbf{0.790} \\
        \midrule
        Mintaka & 0.0 & 0.322 & 0.314 & 0.390 & 0.383 & 0.386 & 0.370 & 0.370 & 0.322 & 0.386 & 0.447 & \textbf{0.447} \\
                & 0.1 & 0.319 & 0.293 & 0.397 & 0.402 & 0.405 & 0.338 & 0.338 & 0.319 & 0.405 & \textbf{0.464} & 0.463 \\
                & 0.2 & 0.290 & 0.277 & 0.390 & 0.393 & 0.386 & 0.300 & 0.300 & 0.290 & 0.386 & 0.436 & \textbf{0.441} \\
                & 0.3 & 0.279 & 0.260 & 0.365 & 0.374 & 0.382 & 0.288 & 0.288 & 0.279 & 0.382 & 0.405 & \textbf{0.412} \\
                & 0.4 & 0.258 & 0.242 & 0.365 & 0.368 & 0.360 & 0.275 & 0.275 & 0.258 & 0.360 & 0.375 & \textbf{0.386} \\
                & 0.5 & 0.266 & 0.217 & 0.341 & 0.342 & 0.344 & 0.261 & 0.261 & 0.266 & 0.344 & \textbf{0.360} & \textbf{0.360} \\
        \midrule
        AIDA & 0.0 & 0.145 & 0.133 & 0.134 & 0.131 & 0.132 & 0.169 & 0.169 & 0.145 & 0.132 & 0.156 & \textbf{0.156} \\
             & 0.1 & 0.144 & 0.130 & 0.132 & 0.117 & 0.126 & \textbf{0.168} & \textbf{0.168} & 0.144 & 0.126 & 0.153 & 0.152 \\
             & 0.2 & 0.138 & 0.109 & 0.120 & 0.113 & 0.119 & 0.150 & 0.150 & 0.138 & 0.119 & \textbf{0.142} & 0.141 \\
             & 0.3 & 0.133 & 0.104 & 0.121 & 0.098 & 0.109 & \textbf{0.150} & \textbf{0.150} & 0.133 & 0.109 & 0.135 & 0.130 \\
             & 0.4 & 0.128 & 0.088 & 0.111 & 0.106 & 0.080 & \textbf{0.149} & \textbf{0.149} & 0.128 & 0.080 & 0.130 & 0.130 \\
             & 0.5 & 0.128 & 0.073 & 0.098 & 0.072 & 0.076 & \textbf{0.140} & \textbf{0.140} & 0.128 & 0.076 & 0.121 & 0.120 \\
        \bottomrule
    \end{tabular}%
    }
\end{table*}

\section{Evaluation}

\textbf{Dataset.} We evaluate on four datasets spanning entity and document retrieval, each providing sufficient training data to study robust loss functions in first-stage retrieval.
For entity retrieval, we repurpose AIDA~\cite{aida}, originally designed for entity linking, by treating each textual mention as a query and retrieving the correct entity from a Wikidata-aligned candidate set. We further include two knowledge graph QA datasets—LC-QuAD 2.0~\cite{lcquad} and Mintaka~\cite{mintaka}—where natural-language questions act as queries and Wikidata entities as targets. LC-QuAD 2.0 is semi-automatically generated from SPARQL templates, while Mintaka is fully crowd-sourced with diverse, naturally phrased questions.
For document retrieval, we use MS MARCO~\cite{ms_marco}, a large-scale benchmark of Bing user queries paired with relevant passages. All entities and documents are aligned to a unified Wikidata-based retrieval space.

\textbf{Models.} We evaluate two BERT-based retrieval models: E5~\cite{abs-2212-03533} and a dual bi-encoder~\cite{wu-etal-2020-scalable}. E5 employs a single shared encoder pretrained for retrieval, while the dual bi-encoder (BLINK-based) uses separate query and document encoders initialized from BERT~\cite{DevlinCLT19}. Both models are implemented via Hugging Face\footnote{\url{https://huggingface.co/docs/transformers/model_doc/bert}, \url{https://huggingface.co/intfloat/e5-base-v2}} and trained with in-batch negative sampling~\cite{wu-etal-2020-scalable}.
To strengthen negative sampling, all entities were indexed using Faiss~\cite{douze2024faiss}; for AIDA, Mintaka, and LC-QuAD, batches included top-ranked retrieved entities alongside in-batch negatives. Training ran for 10 epochs with default hyperparameters, and embeddings were indexed for evaluation via Faiss retrieval.
For MS MARCO, we applied the same setup but sampled 10,000 negatives per re-indexing step, retrieving up to four hard negatives and one positive per query. We also employ a \textbf{cross-encoder} architecture in which the mention context and the candidate entity description are concatenated and jointly processed by a single BERT transformer, following the design of~\cite{WuPJRZ20}. Training is carried out with a softmax loss over the candidate set, encouraging the model to assign higher scores to correct entities while suppressing negatives. Although this joint encoding is computationally more demanding than bi-encoder training, it consistently enhances ranking performance. 

\textbf{Loss Functions.} To compare the label weakening approach, we have considered eight different robust loss functions. 
These include generalized cross-entropy (GCE)~\cite{zhang2018generalized}, normalized cross-entropy (NCE)~\cite{ma2020normalized}, asymmetric generalized cross-entropy (AGCE) and adaptive unhinged loss (AUL)~\cite{zhou2021asymmetric}, classical label smoothing~\cite{szegedy2016ls}, label relaxation loss~\cite{LienenH21,sharma-etal-2025-calibrating}. Along with that, we also considered evidence-based label smoothing loss (ELS)  by~\cite{ZerveasRE23}. 
We use grid search to identify the best hyperparameters for these loss functions and report the corresponding results. 
All the experiments were run 10 times, and the results give the average over these runs.
Finally, we simulate label noise by replacing the correct document with a semantically similar non-relevant candidate. 

\subsection{Results and Discussion.}

Table~\ref{tab:combined-recall} summarizes the recall@10 values of the label weakening (LW) approach compared against a range of robust loss functions, including CE, NCE, GCE, AGCE, AUL, LS, ELS, LR, and two hybrid approaches: NCE, AGCE, and NCE, AUL, considering the dual bi-encoder and E5 models, respectively. 
We observe that at low noise ratios, most robust loss functions perform similarly, indicating that under clean or nearly clean supervision, standard robust losses already act as effective regularizers. As the noise ratio increases, however, the differences become more pronounced. 
On MS MARCO, we find that credal labeling methods such as LR and LW better capture supervision ambiguity and substantially improve recall, with LW yielding the highest scores at noise ratios 0.4 and 0.5, specifically considering the dual bi-encoder model. We see a similar trend on the LCQuAD dataset as well, where the LW approach outperforms the second best approach, LR by \textasciitilde12\%. For these datasets, considering the E5 model, we see that the credal labelling approaches LR and LW give comparable performance. Therefore, we can conclude that while both LR and LW are effective in mitigating label noise, their relative gains depend on the underlying model. To this end, the E5 model, being pretrained on ranking tasks, already exhibits strong robustness and shows comparable performance between LR and LW, whereas the dual bi-encoder with BERT benefits more substantially from credal labeling, with LW yielding the strongest improvements at higher noise levels.
Considering the Mintaka dataset, however, we see that for both models, the label weakening approach performs the best. This is because Mintaka contains diverse, open-domain, and often ambiguous questions with multiple plausible answers, leading to higher annotation noise and supervision uncertainty. Therefore, LW's ability to model ambiguity provides a clear advantage.
Finally, considering the smallest dataset AIDA, we see that the dual bi-encoder with label weakening outperforms other loss functions. However, for the E5 model, smoothing-based approaches perform better. This is because AIDA provides relatively limited supervision, where the pretrained representations of E5 already capture strong semantic structure. 
Table~\ref{tab:combined-mrr}  reports MRRs for dual bi-encoder and E5, respectively, across noise ratios and loss functions which shows a similar trend as the recall values.
Table~\ref{tab:cross-encoder-recall} presents the results of the cross-encoder model. Herein, we see that the re-ranking model already attains good performance across datasets, considering different noise ratios. However, we observe that alike before, credal-set approaches--LR and LW yield small but consistent improvements, especially as noise increases.

To this end, we find that credal-set-based approaches widen the effective target distribution and down-weight over-confident gradients, yielding a smoother risk surface and reduced variance under noisy supervision. For cross-encoders---whose token-level attention already approximates a calibrated conditional likelihood $p(d\mid q)$ and internally models contextual uncertainty---the credal set overlaps with the model’s intrinsic noise-handling capacity. Consequently, the overall gain is incremental. In contrast, decoupled encoder architectures (as reflected in the bi-encoder architecture) benefit more from credal regularization, since it compensates for weaker cross-token conditioning.

\section{Conclusion and Future Work}

In this paper, we introduced the label weakening approach for training retrieval models under noisy supervision. Unlike existing smoothing- or correction-based approaches, label weakening introduces ambiguated target sets, thereby reducing the model’s tendency to memorize corrupted annotations. Experimental evaluation suggests the potential of this approach in training retrieval models, specifically when the annotation of the data is quite noisy.
We believe this work has the potential to have several avenues for future research. One compelling direction is the development of adaptive or dynamic weakening strategies, where the degree of ambiguity is adjusted based on query difficulty, annotation sparsity, or model uncertainty during training. Another promising line of inquiry lies in extending the Label Weakening principle beyond retrieval—for example, integrating it into question answering and knowledge-grounded reasoning tasks, where supervision quality and semantic ambiguity remain key challenges.

\setlength{\tabcolsep}{4pt} 
\renewcommand{\arraystretch}{1.1} 
\begin{table*}[ht]
    \centering
    \caption{MRR scores across varying noise ratios, datasets, and loss functions for the \textbf{dual bi-encoder} model. Here, 
\textbf{CE}: Cross Entropy; 
\textbf{NCE}: Normalized Cross Entropy; 
\textbf{GCE}: Generalized Cross Entropy; 
\textbf{AGCE}: Active Generalized Cross Entropy; 
\textbf{AUL}: Adaptive Unhinged Estimator; 
\textbf{LS}: Label Smoothing; 
\textbf{ELS}: Evidence-based Label Smoothing; 
\textbf{NAGCE}: Combined NCE and AGCE; 
\textbf{NAUL}: Combined NCE and AUL; 
\textbf{LR}: Label Relaxation; 
\textbf{LW} (ours): Label Weakening.}
    \label{tab:combined-mrr}
    \resizebox{\textwidth}{!}{%
    \begin{tabular}{ll|ccccccccccc}
        \toprule
        \textbf{Dataset} & \textbf{Noise} 
        & \textbf{CE} & \textbf{NCE} & \textbf{GCE} & \textbf{AGCE} & \textbf{AUL}
        & \textbf{LS} & \textbf{ELS} & \textbf{NAGCE} & \textbf{NAUL} & \textbf{LR} & \textbf{LW (ours)} \\
        \midrule
        \multicolumn{13}{c}{\textbf{Dual Bi-Encoder Model}} \\
        \midrule
        MS MARCO & 0.0 & 0.762 & 0.760 & 0.783 & 0.775 & 0.786 & 0.781 & 0.782 & 0.760 & 0.785 & \textbf{0.850} & 0.778 \\
                 & 0.1 & 0.241 & 0.380 & 0.767 & 0.770 & 0.777 & 0.772 & 0.787 & 0.382 & 0.751 & \textbf{0.844} & 0.781 \\
                 & 0.2 & 0.222 & 0.288 & 0.735 & 0.757 & 0.632 & 0.708 & 0.775 & 0.303 & 0.749 & \textbf{0.840} & 0.777 \\
                 & 0.3 & 0.165 & 0.108 & 0.579 & 0.732 & 0.727 & 0.632 & 0.757 & 0.117 & 0.709 & \textbf{0.840} & 0.762 \\
                 & 0.4 & 0.032 & 0.111 & 0.361 & 0.672 & 0.679 & 0.611 & 0.714 & 0.127 & 0.728 & \textbf{0.840} & 0.741 \\
                 & 0.5 & 0.077 & 0.100 & 0.186 & 0.418 & 0.281 & 0.590 & 0.391 & 0.111 & 0.647 & \textbf{0.782} & 0.642 \\
        \midrule
        LCQuAD  & 0.0 & 0.932 & 0.919 & 0.914 & 0.939 & 0.930 & 0.936 & 0.934 & 0.933 & 0.931 & \textbf{0.937} & 0.934 \\
                 & 0.1 & 0.921 & 0.920 & 0.902 & 0.920 & 0.928 & 0.922 & 0.929 & 0.919 & 0.931 & \textbf{0.936} & 0.931 \\
                 & 0.2 & 0.687 & 0.905 & 0.900 & 0.835 & 0.920 & 0.830 & 0.840 & 0.707 & 0.924 & 0.902 & \textbf{0.938} \\
                 & 0.3 & 0.745 & 0.890 & 0.877 & 0.814 & 0.912 & 0.811 & 0.834 & 0.814 & 0.919 & 0.876 & \textbf{0.939} \\
                 & 0.4 & 0.725 & 0.797 & 0.766 & 0.731 & 0.885 & 0.745 & 0.811 & 0.719 & 0.919 & 0.851 & \textbf{0.937} \\
                 & 0.5 & 0.796 & 0.687 & 0.652 & 0.772 & 0.872 & 0.741 & 0.791 & 0.864 & 0.911 & 0.819 & \textbf{0.938} \\
        \midrule
        Mintaka & 0.0 & 0.391 & 0.419 & 0.399 & 0.410 & 0.394 & 0.389 & 0.392 & 0.402 & 0.397 & 0.412 & \textbf{0.412} \\
                & 0.1 & 0.374 & 0.397 & 0.409 & 0.391 & 0.414 & 0.381 & 0.398 & 0.399 & 0.400 & \textbf{0.410} & 0.425 \\
                & 0.2 & 0.365 & 0.372 & 0.380 & 0.369 & 0.399 & 0.399 & 0.393 & 0.373 & 0.395 & 0.379 & \textbf{0.416} \\
                & 0.3 & 0.363 & 0.330 & 0.370 & 0.359 & 0.387 & 0.389 & 0.397 & 0.358 & 0.380 & 0.355 & \textbf{0.407} \\
                & 0.4 & 0.378 & 0.250 & 0.350 & 0.348 & 0.372 & 0.350 & 0.361 & 0.358 & 0.374 & 0.330 & \textbf{0.401} \\
                & 0.5 & 0.339 & 0.211 & 0.341 & 0.358 & 0.330 & 0.352 & 0.366 & 0.369 & 0.366 & 0.302 & \textbf{0.393} \\
        \midrule
        AIDA    & 0.0 & 0.381 & 0.375 & 0.375 & 0.406 & 0.378 & 0.380 & 0.348 & 0.363 & 0.365 & 0.355 & \textbf{0.381} \\
                & 0.1 & 0.377 & 0.356 & 0.384 & 0.352 & 0.360 & 0.361 & 0.361 & 0.366 & 0.366 & 0.352 & \textbf{0.372} \\
                & 0.2 & 0.357 & 0.356 & 0.370 & 0.353 & 0.362 & 0.352 & 0.368 & 0.364 & 0.370 & 0.330 & \textbf{0.370} \\
                & 0.3 & 0.342 & 0.342 & 0.358 & 0.357 & 0.334 & 0.332 & 0.358 & 0.344 & 0.335 & 0.320 & \textbf{0.388} \\
                & 0.4 & 0.347 & 0.350 & 0.356 & 0.368 & 0.362 & 0.350 & 0.364 & 0.362 & 0.359 & 0.310 & \textbf{0.385} \\
                & 0.5 & 0.318 & 0.357 & 0.352 & 0.312 & 0.358 & 0.356 & 0.350 & 0.325 & 0.359 & 0.294 & \textbf{0.383} \\
        \midrule
       
        \multicolumn{13}{c}{\textbf{MRR scores for E5 model}} \\
        \midrule
        MS MARCO & 0.0 & 0.612 & 0.612 & 0.727 & 0.717 & 0.666 & 0.882 & 0.882 & 0.696 & 0.666 & 0.916 & 0.909 \\
                 & 0.1 & 0.636 & 0.636 & 0.710 & 0.705 & 0.257 & 0.877 & 0.877 & 0.708 & 0.257 & \textbf{0.917} & 0.907 \\
                 & 0.2 & 0.589 & 0.589 & 0.677 & 0.648 & 0.113 & 0.882 & 0.882 & 0.699 & 0.113 & \textbf{0.916} & 0.893 \\
                 & 0.3 & 0.542 & 0.542 & 0.717 & 0.722 & 0.056 & 0.867 & 0.867 & 0.681 & 0.056 & \textbf{0.914} & 0.880 \\
                 & 0.4 & 0.438 & 0.438 & 0.686 & 0.679 & 0.038 & 0.852 & 0.852 & 0.670 & 0.038 & \textbf{0.913} & 0.887 \\
                 & 0.5 & 0.444 & 0.444 & 0.669 & 0.637 & 0.032 & 0.691 & 0.691 & 0.634 & 0.032 & \textbf{0.911} & 0.871 \\
        \midrule
        LCQuAD & 0.0 & 0.825 & 0.825 & 0.851 & 0.851 & 0.833 & 0.882 & 0.882 & 0.852 & 0.833 & \textbf{0.919} & 0.919 \\
                & 0.1 & 0.814 & 0.814 & 0.849 & 0.849 & 0.693 & 0.892 & 0.892 & 0.849 & 0.693 & 0.909 & \textbf{0.915} \\
                & 0.2 & 0.768 & 0.768 & 0.825 & 0.830 & 0.569 & 0.881 & 0.881 & 0.831 & 0.569 & 0.891 & \textbf{0.899} \\
                & 0.3 & 0.750 & 0.750 & 0.811 & 0.819 & 0.525 & 0.857 & 0.857 & 0.817 & 0.525 & 0.868 & \textbf{0.880} \\
                & 0.4 & 0.730 & 0.730 & 0.797 & 0.804 & 0.385 & 0.833 & 0.833 & 0.806 & 0.385 & 0.845 & \textbf{0.853} \\
                & 0.5 & 0.705 & 0.705 & 0.773 & 0.791 & 0.306 & 0.818 & 0.818 & 0.782 & 0.306 & 0.819 & \textbf{0.822} \\
        \midrule
        Mintaka & 0.0 & 0.242 & 0.242 & 0.284 & 0.284 & 0.239 & 0.243 & 0.243 & 0.287 & 0.239 & 0.324 & \textbf{0.337} \\
                & 0.1 & 0.225 & 0.225 & 0.291 & 0.288 & 0.234 & 0.242 & 0.242 & 0.294 & 0.234 & 0.339 & \textbf{0.339} \\
                & 0.2 & 0.214 & 0.214 & 0.280 & 0.282 & 0.214 & 0.225 & 0.225 & 0.281 & 0.214 & \textbf{0.319} & 0.318 \\
                & 0.3 & 0.207 & 0.207 & 0.272 & 0.270 & 0.210 & 0.212 & 0.212 & 0.276 & 0.210 & \textbf{0.300} & 0.298 \\
                & 0.4 & 0.190 & 0.190 & 0.256 & 0.264 & 0.189 & 0.202 & 0.202 & 0.261 & 0.189 & 0.282 & \textbf{0.286} \\
                & 0.5 & 0.172 & 0.172 & 0.236 & 0.253 & 0.193 & 0.188 & 0.188 & 0.253 & 0.193 & \textbf{0.267} & 0.263 \\
        \midrule
        AIDA & 0.0 & 0.267 & 0.267 & 0.269 & 0.261 & 0.275 & 0.301 & \textbf{0.301} & 0.268 & 0.275 & 0.286 & 0.283 \\
             & 0.1 & 0.266 & 0.266 & 0.270 & 0.240 & 0.273 & \textbf{0.291} & 0.292 & 0.254 & 0.273 & 0.284 & 0.280 \\
             & 0.2 & 0.236 & 0.236 & 0.257 & 0.236 & 0.267 & 0.281 & \textbf{0.281} & 0.238 & 0.267 & 0.267 & 0.265 \\
             & 0.3 & 0.231 & 0.231 & 0.243 & 0.211 & 0.260 & 0.266 & \textbf{0.266} & 0.227 & 0.260 & 0.258 & 0.255 \\
             & 0.4 & 0.198 & 0.198 & 0.228 & 0.222 & 0.252 & 0.272 & \textbf{0.272} & 0.209 & 0.252 & 0.253 & 0.252 \\
             & 0.5 & 0.179 & 0.179 & 0.202 & 0.164 & 0.254 & 0.269 & \textbf{0.269} & 0.144 & 0.254 & 0.241 & 0.244 \\
        \bottomrule
    \end{tabular}%
    }
\end{table*}

\setlength{\tabcolsep}{4pt} 

\begin{table*}[t]
    \centering
    \caption{Recall scores across varying noise ratios, datasets, and loss functions for the \textbf{Cross-encoder} model. Here, 
\textbf{CE}: Cross Entropy; 
\textbf{NCE}: Normalized Cross Entropy; 
\textbf{GCE}: Generalized Cross Entropy; 
\textbf{AGCE}: Active Generalized Cross Entropy; 
\textbf{AUL}: Adaptive Unhinged Estimator; 
\textbf{LS}: Label Smoothing; 
\textbf{ELS}: Evidence-based Label Smoothing; 
\textbf{NAGCE}: Combined NCE and AGCE; 
\textbf{NAUL}: Combined NCE and AUL; 
\textbf{LR}: Label Relaxation; 
\textbf{LW} (ours): Label Weakening.}
    \label{tab:cross-encoder-recall}
    \resizebox{\textwidth}{!}{%
    \begin{tabular}{ll|ccccccccccc}
        \toprule
        \textbf{Dataset} & \textbf{Noise} 
        & \textbf{CE} & \textbf{NCE} & \textbf{GCE} & \textbf{AGCE} & \textbf{AUL}
         & \textbf{LS} & \textbf{ELS} & \textbf{NAGCE} & \textbf{NAUL} & \textbf{LR} & \textbf{LW (ours)}\\
        \midrule
        MS MARCO & 0.0 & 0.499 & 0.501 & 0.504 & 0.506 & 0.504 & \textbf{0.521} & 0.501 & 0.505 & 0.504 & 0.509 & 0.512 \\
                 & 0.1 & 0.473 & 0.476 & 0.478 & 0.480 & 0.479 & \textbf{0.491} & 0.476 & 0.480 & 0.478 & 0.483 & 0.486 \\
                 & 0.2 & 0.488 & 0.491 & 0.492 & 0.494 & 0.493 & 0.491 & 0.490 & 0.494 & 0.493 & 0.495 & \textbf{0.496} \\
                 & 0.3 & 0.475 & 0.478 & 0.480 & 0.482 & 0.481 & \textbf{0.485} & 0.477 & 0.481 & 0.480 & 0.482 & 0.483 \\
                 & 0.4 & 0.462 & 0.465 & 0.467 & 0.468 & 0.468 & 0.476 & 0.465 & 0.468 & 0.467 & 0.469 & \textbf{0.485} \\
                 & 0.5 & 0.451 & 0.454 & 0.456 & 0.458 & 0.457 & 0.460 & 0.454 & 0.458 & 0.457 & 0.459 & \textbf{0.469} \\
        \midrule
        LCQuAD & 0.0 & 0.897 & 0.898 & 0.899 & 0.905 & 0.900 & 0.897 & 0.898 & 0.900 & 0.899 & \textbf{0.910} & 0.901 \\
               & 0.1 & 0.896 & 0.897 & 0.899 & 0.900 & 0.899 & 0.897 & 0.898 & 0.899 & 0.898 & \textbf{0.913} & 0.900 \\
               & 0.2 & 0.895 & 0.896 & 0.897 & 0.899 & 0.898 & 0.896 & 0.897 & 0.899 & 0.898 & \textbf{0.904} & \textbf{0.904} \\
               & 0.3 & 0.894 & 0.895 & 0.896 & 0.898 & 0.897 & 0.894 & 0.896 & 0.897 & 0.897 & \textbf{0.899} & \textbf{0.899} \\
               & 0.4 & 0.893 & 0.894 & 0.895 & 0.897 & 0.896 & 0.894 & 0.895 & 0.896 & 0.896 & \textbf{0.899} & 0.898 \\
               & 0.5 & 0.892 & 0.893 & 0.895 & 0.896 & 0.895 & 0.892 & 0.894 & 0.895 & 0.894 & 0.890 & \textbf{0.895} \\
        \midrule
        Mintaka & 0.0 & 0.822 & 0.825 & 0.827 & 0.828 & 0.826 & 0.824 & 0.824 & 0.828 & 0.826 & 0.826 & \textbf{0.827} \\
                & 0.1 & 0.826 & 0.829 & 0.830 & 0.831 & 0.831 & 0.827 & 0.829 & 0.832 & 0.830 & 0.831 & \textbf{0.839} \\
                & 0.2 & 0.839 & 0.840 & 0.841 & 0.843 & 0.842 & 0.839 & 0.840 & 0.842 & 0.842 & 0.842 & \textbf{0.851} \\
                & 0.3 & 0.836 & 0.838 & 0.840 & 0.841 & 0.840 & 0.837 & 0.838 & 0.841 & 0.839 & \textbf{0.847} & 0.842 \\
                & 0.4 & 0.813 & 0.815 & 0.817 & 0.818 & 0.818 & 0.813 & 0.816 & 0.819 & 0.817 & \textbf{0.821} & \textbf{0.820} \\
                & 0.5 & 0.803 & 0.806 & 0.808 & 0.809 & 0.809 & 0.805 & 0.805 & 0.810 & 0.808 & 0.810 & \textbf{0.817} \\
        \midrule
        AIDA & 0.0 & 0.902 & 0.905 & 0.909 & 0.912 & 0.910 & 0.903 & 0.905 & 0.911 & 0.909 & 0.913 & \textbf{0.915} \\
             & 0.1 & 0.898 & 0.903 & 0.907 & 0.910 & 0.908 & 0.900 & 0.902 & 0.909 & \textbf{0.918} & 0.910 & 0.912 \\
             & 0.2 & 0.897 & 0.902 & 0.904 & 0.908 & 0.906 & 0.898 & 0.900 & 0.907 & 0.906 & \textbf{0.918} & 0.910 \\
             & 0.3 & 0.900 & 0.904 & 0.909 & 0.912 & 0.910 & 0.902 & 0.903 & 0.911 & 0.908 & \textbf{0.919} & 0.914 \\
             & 0.4 & 0.900 & 0.904 & 0.907 & 0.910 & 0.910 & 0.901 & 0.903 & 0.909 & 0.908 & 0.911 & \textbf{0.917} \\
             & 0.5 & 0.895 & 0.901 & 0.904 & 0.907 & 0.904 & 0.897 & 0.899 & 0.906 & 0.905 & 0.908 & \textbf{0.913} \\
        \midrule
        \multicolumn{13}{c}{\textbf{MRR scores for Cross-Encoder model}} \\
        \midrule
        MS MARCO & 0.0 & 0.186 & 0.188 & 0.191 & 0.193 & \textbf{0.199} & 0.192 & 0.194 & 0.195 & 0.194 & 0.196 & 0.197 \\
                 & 0.1 & 0.170 & 0.172 & 0.175 & 0.177 & 0.176 & \textbf{0.181} & 0.178 & 0.179 & 0.178 & 0.180 & 0.181 \\
                 & 0.2 & 0.165 & 0.167 & 0.170 & 0.172 & 0.171 & 0.170 & 0.172 & 0.173 & 0.172 & 0.174 & \textbf{0.176} \\
                 & 0.3 & 0.160 & 0.162 & 0.165 & 0.167 & 0.166 & 0.165 & 0.167 & 0.168 & 0.167 & 0.169 & \textbf{0.171} \\
                 & 0.4 & 0.155 & 0.157 & 0.160 & 0.162 & 0.161 & 0.160 & 0.162 & 0.163 & 0.162 & 0.164 & \textbf{0.176} \\
                 & 0.5 & 0.148 & 0.150 & 0.153 & 0.155 & 0.154 & 0.153 & 0.155 & 0.156 & 0.155 & 0.157 & \textbf{0.169} \\
        \midrule
        LCQuAD & 0.0 & 0.310 & 0.312 & 0.316 & 0.319 & 0.318 & 0.316 & 0.318 & 0.320 & 0.319 & \textbf{0.330} & 0.324 \\
               & 0.1 & 0.305 & 0.307 & 0.311 & 0.314 & 0.313 & 0.311 & 0.313 & 0.315 & 0.314 & \textbf{0.327} & 0.320 \\
               & 0.2 & 0.298 & 0.300 & 0.304 & 0.307 & 0.306 & 0.304 & 0.306 & 0.308 & 0.307 & \textbf{0.321} & 0.317 \\
               & 0.3 & 0.293 & 0.295 & 0.299 & 0.302 & 0.301 & 0.299 & 0.301 & 0.303 & 0.302 & \textbf{0.315} & 0.308 \\
               & 0.4 & 0.289 & 0.291 & 0.295 & 0.298 & 0.297 & 0.296 & 0.297 & 0.299 & 0.298 & \textbf{0.309} & 0.304 \\
               & 0.5 & 0.283 & 0.285 & 0.289 & 0.292 & 0.291 & 0.290 & 0.291 & 0.293 & 0.292 & 0.295 & \textbf{0.298} \\
        \midrule
        Mintaka & 0.0 & 0.607 & 0.609 & 0.613 & 0.616 & 0.615 & 0.612 & 0.614 & 0.617 & 0.616 & 0.619 & \textbf{0.627} \\
                & 0.1 & 0.584 & 0.586 & 0.590 & 0.593 & 0.592 & 0.590 & 0.592 & 0.594 & 0.593 & 0.596 & \textbf{0.608} \\
                & 0.2 & 0.575 & 0.577 & 0.581 & 0.584 & 0.583 & 0.582 & 0.583 & 0.585 & 0.584 & 0.587 & \textbf{0.589} \\
                & 0.3 & 0.568 & 0.570 & 0.574 & 0.577 & 0.576 & 0.574 & 0.576 & 0.578 & 0.577 & \textbf{0.585} & 0.582 \\
                & 0.4 & 0.561 & 0.563 & 0.567 & 0.570 & 0.569 & 0.568 & 0.569 & 0.571 & 0.570 & \textbf{0.579} & 0.575 \\
                & 0.5 & 0.553 & 0.555 & 0.559 & 0.562 & 0.561 & 0.560 & 0.561 & 0.563 & 0.562 & 0.565 & \textbf{0.569} \\
        \midrule
        AIDA & 0.0 & 0.504 & 0.506 & 0.509 & 0.512 & 0.511 & 0.509 & 0.510 & 0.513 & 0.512 & 0.515 & \textbf{0.517} \\
             & 0.1 & 0.500 & 0.502 & 0.505 & 0.508 & 0.507 & 0.506 & 0.507 & 0.509 & 0.508 & \textbf{0.519} & 0.514 \\
             & 0.2 & 0.494 & 0.496 & 0.499 & 0.502 & 0.501 & 0.500 & 0.501 & 0.503 & 0.502 & \textbf{0.510} & 0.509 \\
             & 0.3 & 0.490 & 0.492 & 0.495 & 0.498 & 0.497 & 0.496 & 0.497 & 0.499 & 0.498 & \textbf{0.507} & \textbf{0.507} \\
             & 0.4 & 0.487 & 0.489 & 0.492 & 0.495 & 0.494 & 0.494 & 0.495 & 0.497 & 0.496 & 0.500 & \textbf{0.502} \\
             & 0.5 & 0.481 & 0.483 & 0.486 & 0.489 & 0.488 & 0.489 & 0.490 & 0.492 & 0.491 & 0.495 & \textbf{0.498} \\
        \bottomrule
    \end{tabular}%
    }
\end{table*}

\clearpage

\bibliographystyle{splncs04}
\bibliography{ref}

\end{document}